# Detecting HIV-Related Stigma in Clinical Narratives Using Large Language Models

**Ziyi Chen, MS[1], Yasir Khan, MS[1], Mengyuan Zhang, MS[1], Cheng Peng, PhD[1], Mengxian Lyu, MS[1], Yiyang Liu, PhD[2], Krishna Vaddiparti, PhD[2], Robert L Cook, PhD[2], Mattia Prosperi, PhD[2], Yonghui Wu, PhD[1,3*]**

**[1]Department of Health Outcomes and Biomedical Informatics, College of Medicine, University of Florida, Gainesville, FL, USA; [2]Department of Epidemiology, College of Public Health and Health Professions, University of Florida, Gainesville, FL, USA; [3]Preston A. Wells, Jr. Center for Brain Tumor Therapy, Lillian S. Wells Department of Neurosurgery, University of Florida, Gainesville, FL, USA**

**Abstract**

*Human immunodeficiency virus (HIV)-related stigma is a critical psychosocial determinant of health for people living with HIV (PLWH), influencing mental health, engagement in care, and treatment outcomes. Although stigma-related experiences are documented in clinical narratives, there is a lack of off-the-shelf tools to extract and categorize them. This study aims to develop a large language model (LLM)-based tool for identifying HIV stigma from clinical notes. We identified clinical notes from PLWH receiving care at the University of Florida (UF) Health between 2012 and 2022. Candidate sentences were identified using expert-curated stigma-related keywords and iteratively expanded via clinical word embeddings. A total of 1,332 sentences were manually annotated across four stigma subscales: Concern with Public Attitudes, Disclosure Concerns, Negative Self-Image, and Personalized Stigma. We compared GatorTron-large and BERT as encoder-based baselines, and GPT-OSS-20B, LLaMA-8B, and MedGemma-27B as generative LLMs, under zero-shot and few-shot prompting. GatorTron-large achieved the best overall performance (Micro F1 = 0.62). Few-shot prompting substantially improved generative model performance, with 5-shot GPT-OSS-20B and LLaMA-8B achieving Micro-F1 scores of 0.57 and 0.59, respectively. Performance varied by stigma subscale, with Negative Self-Image showing the highest predictability and Personalized Stigma remaining the most challenging. Zero-shot generative inference exhibited non-trivial failure rates (up to 32%). This study develops the first practical NLP tool for identifying HIV stigma in clinical notes.*

## Introduction

The human immunodeficiency virus (HIV), the etiologic agent of acquired immunodeficiency syndrome (AIDS), continues to pose a substantial global public health challenge. According to the World Health Organization, approximately 40.8 million individuals were living with HIV worldwide at the end of 2024[1]. Beyond its physiological burden, HIV-related stigma remains a pervasive and multifaceted barrier to optimal care for people living with HIV (PLWH). Stigma operates across multiple psychosocial dimensions, including fear of disclosure, internalized negative self-perceptions, concerns about public attitudes, and direct, personalized experiences of discrimination. These interrelated dimensions shape patients' interactions with health systems and social networks[2,3]. Clinically, it might lead to delayed diagnosis, reduced engagement and retention in care, adverse mental health outcomes, and suboptimal adherence to antiretroviral therapy[4–9]. So systematic assessment and targeted mitigation of HIV-related stigma are essential components of comprehensive care for PLWH. They are critical to improving both short- and long-term clinical and psychosocial outcomes.

Traditional approaches to measuring stigma have largely relied on structured surveys and questionnaires, typically using standardized instruments designed for PLWH. For example, the Berger Stigma Scale[10] consists of 40 Likert-scale items that evaluate four core dimensions of HIV-related stigma. Similarly, Holzemer et al.[11] developed and validated a multidimensional instrument to assess perceived stigma, enabling researchers to establish baseline levels, track changes over time, and evaluate progress toward stigma reduction. Although these approaches are well defined within survey-based frameworks, their representation in routine clinical documentation is considerably less structured. In addition, survey-based approaches present practical limitations: they are time- and resource-intensive to administer and can impose a substantial burden on respondents, particularly in longitudinal studies or routine clinical settings.

Electronic health records (EHRs) offer an alternative source for understanding stigma in real-world care settings. Clinical notes contain rich longitudinal narratives documenting patients' lived experiences, emotional states, and social contexts as observed by healthcare providers. Within these narratives, HIV-related stigma may manifest indirectly through descriptions of emotional distress, social isolation, fear of judgment, or reluctance to disclose HIV status. However, extracting such nuanced psychosocial signals from unstructured clinical text remains a challenging problem for natural language processing (NLP). Existing computational research on HIV-related stigma in clinical data remains limited. Prior work has largely focused on topic modeling approaches to characterize stigma-related themes and social circumstances among PLWH[12], while predictive modeling studies in HIV research have primarily addressed related tasks such as predicting HIV care disengagement[13] or HIV risk[14,15] using clinical data elements.

Traditional rule-based or keyword-driven NLP approaches are insufficient for capturing subtle or implicit expressions of stigma due to the well-known generalizability issue. Recent advances in transformer-based language models, particularly large language models (LLMs), offer new opportunities for modeling complex semantic and contextual relationships in clinical text. Domain-adapted transformer models pretrained on EHR corpora have achieved strong performance across a variety of clinical NLP tasks. At the same time, instruction-following LLMs demonstrate the ability to perform classification through prompting alone, even with minimal or no task-specific training data. These developments raise important questions regarding the trade-offs between fine-tuned domain-specific models and inference-based LLM approaches for sensitive clinical applications such as stigma detection.

In this study, we develop an LLM-based HIV stigma extraction system for clinical notes**,** focusing on four theoretically grounded stigma subscales: (1) Concern with Public Attitudes, (2) Disclosure Concerns, (3) Negative Self-Image, and (4) Personalized Stigma, followed by the Berger stigma scale[10]. Using a rigorously curated and expert-annotated dataset derived from clinical narratives of PLWH at the University of Florida (UF) Health, we evaluate multiple LLMs, including (1) fine-tuned, domain-adapted transformer models and (2) large generative LLMs operating under zero-shot and few-shot prompting settings. Beyond standard performance metrics, we introduce failure rate as a reliability measure to assess the practical usability of LLMs for stigma extraction.

Our contributions are threefold. First, we built the first HIV stigma extraction tool from real-world clinical notes annotated by experts across 4 well-defined stigma dimensions. Second, we provide a direct comparison between supervised fine-tuning and prompt-based inference for stigma determination, testing whether few-shot LLMs can achieve performance comparable to traditional full-dataset tuning. Third, we analyze performance variability across stigma subscales, offering insights into how linguistic variations and conceptual complexity influence model performance. To the best of our knowledge, this is the first open-source NLP tool to extract and categorize HIV stigma types from clinical narratives.

## Methods

### Data source

This study uses EHRs from PLWH who received care at UF Health between 2012 and 2022. The study protocol was approved by the University of Florida Institutional Review Board (IRB202300703). Data were extracted from the UF Integrated Data Repository (IDR), a centralized clinical data warehouse comprising fully digitized patient records from inpatient, outpatient, and emergency department settings, sourced from multiple clinical information systems. All protected health information was removed during data quality assurance procedures.

The clinical notes corpus consisted of provider-authored free-text documents, including progress notes, plans of care, admission notes, and discharge summaries. Scanned documents requiring optical character recognition were excluded. PLWH were identified using a previously validated computable phenotype algorithm[16]. Briefly, patients were classified as PLWH if they had at least one HIV diagnosis code and corroborating evidence from HIV-related laboratory results, antiretroviral therapy prescriptions or dispensations, or multiple HIV-related clinical encounters. Manual chart review demonstrated high performance, with a sensitivity of 98.9% and a specificity of 97.6%.

## Data Preparation and Annotation

HIV experts compiled a list of stigma-related keywords based on their domain knowledge and the questions from established HIV stigma measures. Based on this keyword list, we used a snowball strategy to iteratively review a small batch of randomly sampled notes from the full corpus, using a uniform random sampling procedure to identify new keywords until no more could be found. To further expand the keyword list, we also identified synonyms for these keywords by calculating cosine similarity using word embeddings generated by GatorTron[17], a clinical large language model developed at UF. After applying the snowball strategy, extending synonyms via word embeddings, and conducting manual review by clinicians, the final keyword list contains 91 words.

We extracted candidate sentences from 2.9 million clinical notes in the HIV cohort that contained at least one predefined keyword, yielding 779,702 sentences from 474,470 unique notes. Duplicated sentences were removed and then randomly sampled for manual review, stratified by the proportion of keywords present. When fewer than 10 sentences were available for a given keyword set, all sentences were included. In total, 1,332 sentences were selected for annotation. Three reviewers independently annotated each sentence according to standardized annotation guidelines shown in **Appendix Table 1**, assigning one or more labels across four stigma-related domains: 'Concern with Public Attitudes', 'Disclosure Concerns', 'Negative Self-Image', and 'Personalized Stigma'. Two annotators reviewed each sentence to assess inter-annotator agreement. The initial entity kappa score is 0.286. Sentences for which agreement could not be reached were adjudicated by HIV domain experts using project meetings.

Following adjudication, 230 sentences were confirmed as stigma-related and annotated with one or more domain labels. We constructed four separate datasets, one for each stigma domain, each framed as a binary classification task indicating whether the target outcome was present or absent. For each domain, the annotated sentences were split into training, validation, and test sets using a 6:2:2 ratio. Detailed data information is shown in **Table 1.** Across all subscales, negative labels dominate, indicating substantial class imbalance, particularly for 'Concern with Public Attitudes', where only 14.3% of instances are positive. 'Disclosure Concern' exhibits the highest prevalence of positive cases (36.5%). This imbalance motivates the use of macro-averaged metrics at the subscale level and partially explains the divergent precision–recall trade-offs observed across models.

**Table 1.** Distribution Across Annotated HIV Stigma Subscales

| | Sample Size | Yes (%) | No (%) |
|---|---|---|---|
| **Personalized Stigma** | 230 | 61 (26.5%) | 169 (73.5%) |
| **Disclosure Concern** | 230 | 84 (36.5%) | 146 (63.5%) |
| **Negative Self-image** | 230 | 52 (22.6%) | 178 (77.4%) |
| **Concern with Public Attitudes** | 230 | 33 (14.3%) | 197 (85.7%) |

## Model Training and Evaluation

We trained an HIV stigma prediction model using GatorTron-large, an 8.9B 512-token variant of the GatorTron[17] architecture pretrained on a large-scale corpus of University of Florida Health clinical notes. Model fine-tuning was formulated as a binary sentence-level classification task[18], in which the model predicted whether a domain-specific outcome contained stigma-related content ("met") or not ("not met"). In addition to GatorTron-large, we evaluated both encoder-based and decoder-based LLMs for HIV stigma detection. For encoder-based models, we fine-tuned BERT[19], a widely used transformer architecture for clinical and biomedical prediction tasks, using the same supervised training setup. Model parameters were optimized with the F1 score as the primary objective. All supervised models were trained for up to 50 epochs. GatorTron-large and BERT were fine-tuned with a learning rate of $1\times10^{-6}$.

We further assessed the performance of generative LLMs under zero-shot and few-shot prompting paradigms (1-, 3-, and 5-shot). Specifically, we evaluated LLaMA-3.1-8B[20] with an extended context window of up to 8,000 tokens;

GPT-OSS[21], a 20B-parameter open-weighted model optimized for scalable fine-tuning; and MedGemma-27B[22], a domain-specific model for medical and clinical text understanding. For these models, stigma classification was performed via prompt-based inference without task-specific fine-tuning. **Table 2** presents the few-shot prompt template of 'Disclosure Concerns' stigma. The prompt explicitly assigns the model a domain-expert role. It then provides positive-only examples. By framing the task as a binary clinical judgment and constraining the output format, the template aims to reduce generation variance while preserving semantic inference. Model performance for the subcategory was evaluated using macro-F1, accuracy, macro-precision, and macro-recall. Overall performance was assessed using micro-averaged metrics, calculated by summing the true positives (TP), false positives (FP), false negatives (FN), and true negatives (TN) across all predictions from the four subscales and then computing the corresponding evaluation measures. The failure rate was also used to measure the proportion of test instances for which a model failed to produce a valid prediction. The F1 score was used as the primary metric for model comparison. The end-to-end workflow for developing the HIV stigma prediction model is shown in **Figure 1**.

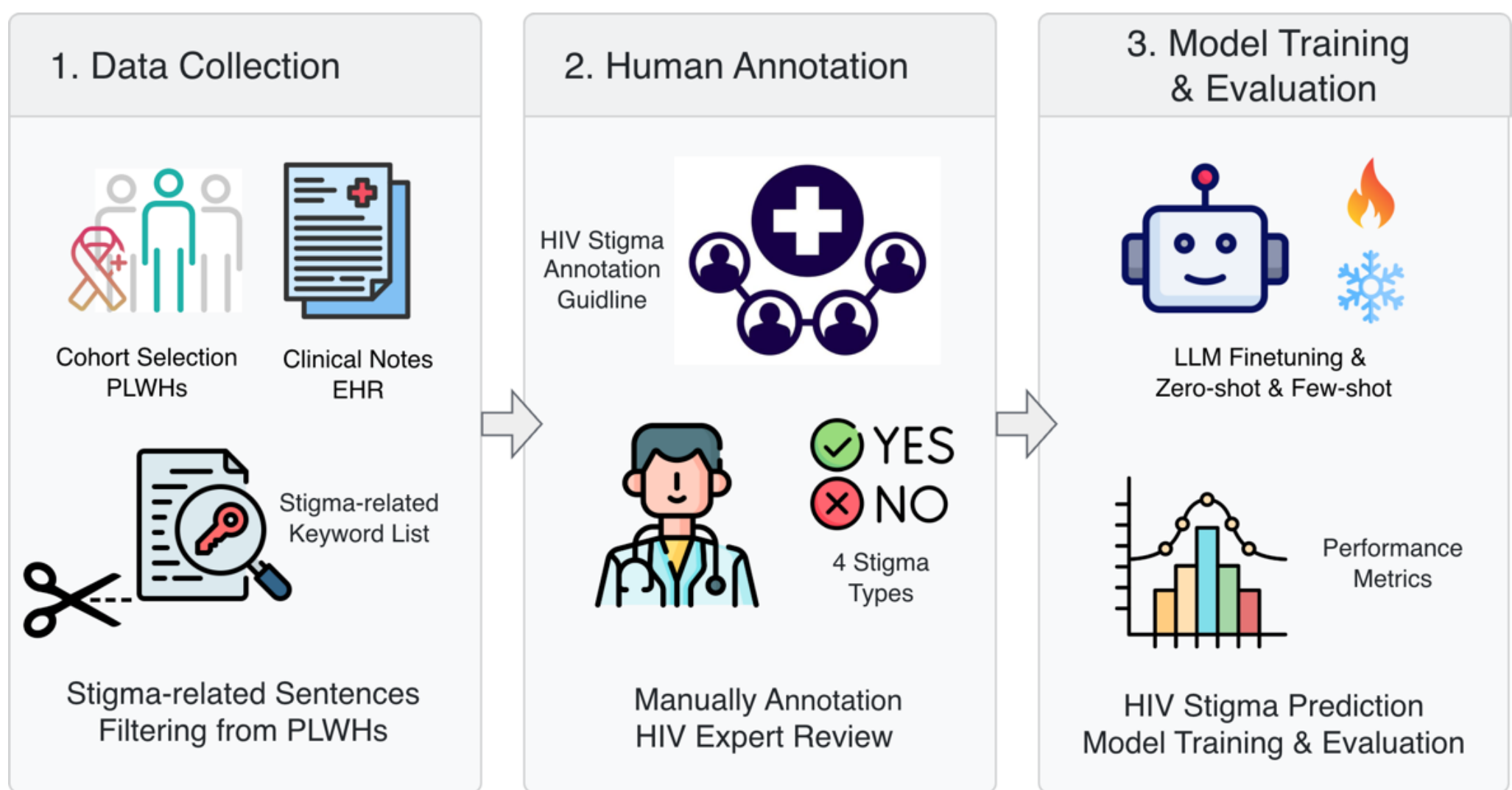

**Figure 1.** End-to-end workflow for developing the HIV stigma prediction model.

**Table 2.** Prompt template of Disclosure Concerns stigma

{"**id**": "xxx",

"**input**": "You are an HIV clinician-scientist with expertise in psychosocial HIV stigma.
Below are examples of sentences that DO express Disclosure Concerns stigma:

Context: She has NOT disclosed her diagnosis to anyone.\nAnswer: yes
Context: he is keeping his diagnosis a secret.\nAnswer: yes
Context: It is an unspoken secret within the family.\nAnswer: yes

Now answer the following question.
Context: [** NAME **] suggested that Pt's secrecy about his HIV status (around family and friends) could be one source of his anxiety
Question: Does this sentence express Disclosure Concerns stigma? Only respond with 'yes' or 'no'.",

"**output**": "yes"}

## Results

**Table 3** and **Figure 2** report overall micro-averaged performance across all stigma dimensions. Among fine-tuned baselines, GT-large achieves the best overall performance (Micro F1 = 0.6190), outperforming BERT fine-tuning (Micro F1 = 0.5455). GT-large demonstrates a notably higher recall (0.5652 vs. 0.4565), indicating better sensitivity to positive stigma instances, while maintaining comparable precision, accuracy, and failure rate. These results suggest that the domain-adapted GT-large architecture is more effective at capturing heterogeneous stigma expressions than a generic BERT fine-tuning baseline.

For large language models evaluated in an inference-only setting, few-shot prompting substantially improves performance across all models. GPT-OSS-20B and LLaMA-8B show consistent gains as the number of shots increases, with peak Micro F1 scores of 0.5745 (GPT-OSS-20B, 5-shot) and 0.5905 (LLaMA-8B, 5-shot), approaching the performance of fine-tuned baselines. Zero-shot performance is remarkably lower, particularly in precision, highlighting the importance of in-context examples for calibration in clinical stigma classification tasks.

Failure rates are low for fine-tuned models and few-shot LLM settings, but non-trivial in zero-shot inference, particularly for LLaMA-8B and MedGemma-27B (up to 32%). This highlights an important practical consideration: while LLMs can approach fine-tuned performance, prompt stability and response reliability depend strongly on minimal in-context supervision.

Performance varies substantially across stigma subscales, reflecting differences in linguistic explicitness and conceptual difficulty, shown in **Figure 3** and **Table 4-7**. GT-large, the best LLM, achieved the best performance for ‘Negative Self-Image’, where few-shot LLMs not only match but often outperform fully fine-tuned models, achieving Macro F1 scores above 0.90 with 3–5 shots. ‘Disclosure Concerns’ also yields relatively high and stable performance across models. In contrast, ‘Personalized Stigma’ remains the most challenging case.

**Table 3.** Overall Micro-Averaged Performance Across HIV Stigma Classification Models

| Model | Methods | Micro F1 | Micro Precision | Micro Recall | Accuracy | Failure Rate |
|---|---|---|---|---|---|---|
| **GT-large** | SFT | ***0.6190*** | ***0.6842*** | **0.5652** | ***0.7895*** | ***0.0000*** |
| BERT | SFT | 0.5455 | 0.6774 | 0.4565 | 0.7697 | 0.0000 |
| GPT-OSS-20b | 0-shot | 0.4554 | 0.4182 | 0.5000 | 0.6382 | 0.0000 |
| | 1-shot | 0.5532 | 0.5417 | 0.5652 | 0.7237 | 0.0000 |
| | 3-shot | 0.5455 | 0.5094 | 0.5870 | 0.7039 | 0.0000 |
| | 5-shot | **0.5745** | **0.5625** | **0.5870** | **0.7368** | 0.0000 |
| LLaMA-8b | 0-shot | 0.4375 | 0.3559 | 0.5676 | 0.5179 | 0.2632 |
| | 1-shot | 0.4946 | 0.4894 | 0.5000 | 0.6908 | 0.0000 |
| | 3-shot | 0.5333 | 0.4746 | 0.6087 | 0.6776 | 0.0000 |
| | 5-shot | **0.5905** | **0.5254** | **0.6739** | **0.7171** | 0.0000 |
| MedGemma-27b | 0-shot | 0.5625 | 0.4355 | ***0.7941*** | 0.6111 | **0.2895** |
| | 1-shot | 0.4091 | 0.4286 | 0.3913 | 0.6579 | 0.0000 |
| | 3-shot | 0.5361 | **0.5098** | 0.5652 | 0.7039 | 0.0000 |
| | 5-shot | **0.5631** | 0.5088 | 0.6304 | **0.7039** | 0.0000 |

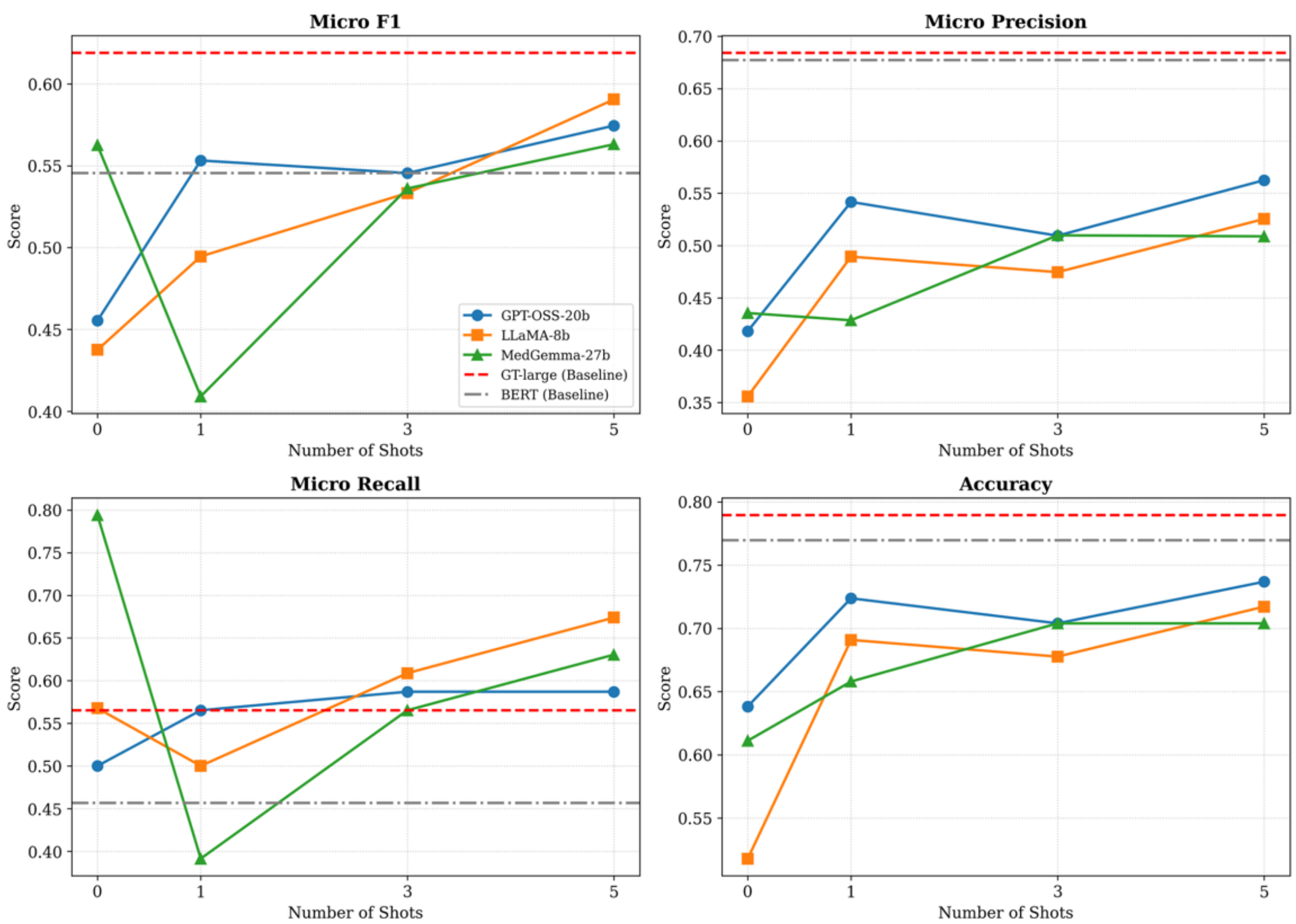

**Figure 2.** Overall Micro-F1, Micro-Precision, Micro-Recall, and Accuracy Performance Across HIV Stigma Classification Models

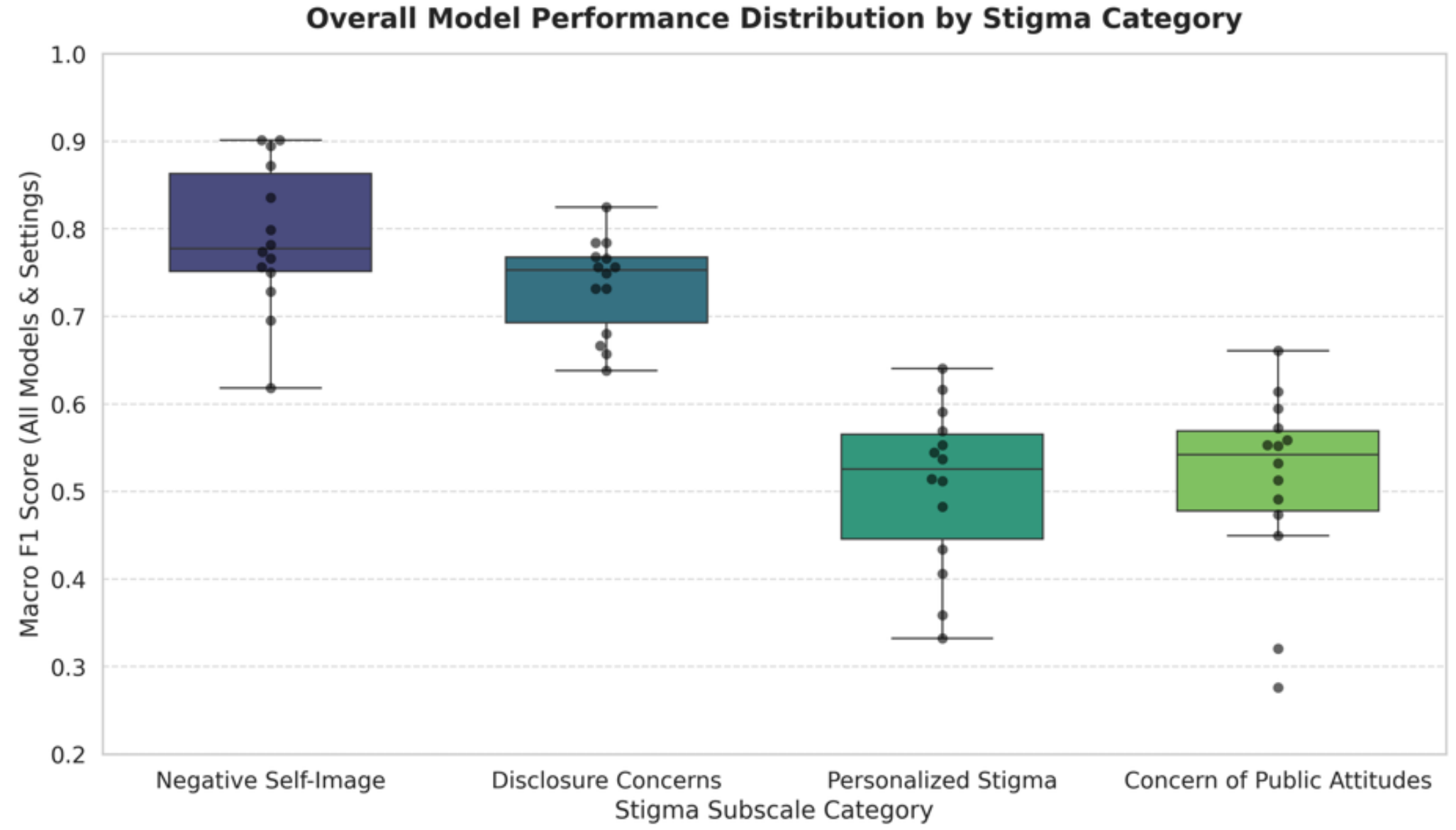

**Figure 3.** Overall Model Performance Distribution by Stigma Subcategory

**Table 4.** Subscale Performance on Concern with Public Attitudes

| Model | Methods | Macro F1 | Macro Precision | Macro Recall | Accuracy | Failure Rate |
|---|---|---|---|---|---|---|
| **GT-large** | SFT | 0.5589 | **0.6667** | 0.5553 | **0.8158** | 0.0000 |
| BERT | SFT | 0.4493 | 0.4079 | 0.5000 | **0.8158** | 0.0000 |
| GPT-OSS-20b | 0-shot | 0.4737 | 0.4872 | 0.4816 | 0.6053 | 0.0000 |
| | 1-shot | 0.5947 | 0.6108 | 0.6797 | 0.6579 | 0.0000 |
| | 3-shot | 0.4911 | 0.5602 | 0.5991 | 0.5263 | 0.0000 |
| | 5-shot | 0.5529 | 0.5889 | 0.6475 | 0.6053 | 0.0000 |
| LLaMA-8b | 0-shot | 0.3203 | 0.3631 | 0.3083 | 0.3846 | **0.3158** |
| | 1-shot | 0.6140 | 0.6090 | 0.6567 | 0.7105 | 0.0000 |
| | 3-shot | 0.5723 | 0.5804 | 0.6244 | 0.6579 | 0.0000 |
| | 5-shot | **0.6607** | 0.6523 | **0.7281** | 0.7368 | 0.0000 |
| MedGemma-27b | 0-shot | 0.2759 | 0.4203 | 0.4203 | 0.2759 | 0.2368 |
| | 1-shot | 0.5522 | 0.5681 | 0.6083 | 0.6316 | 0.0000 |
| | 3-shot | 0.5324 | 0.5568 | 0.5922 | 0.6053 | 0.0000 |
| | 5-shot | 0.5128 | 0.5462 | 0.5760 | 0.5789 | 0.0000 |

**Table 5.** Subscale Performance on Negative Self-Image

| Model | Methods | Macro F1 | Macro Precision | Macro Recall | Accuracy | Failure Rate |
|---|---|---|---|---|---|---|
| **GT-large** | SFT | 0.8355 | 0.8464 | 0.8266 | 0.8684 | 0.0000 |
| BERT | SFT | 0.7662 | 0.7565 | 0.7980 | 0.7895 | 0.0000 |
| GPT-OSS-20b | 0-shot | 0.7564 | 0.7477 | 0.7710 | 0.7895 | 0.0000 |
| | 1-shot | 0.7816 | 0.8479 | 0.7542 | 0.8421 | 0.0000 |
| | 3-shot | 0.8721 | 0.8721 | 0.8721 | 0.8947 | 0.0000 |
| | 5-shot | 0.8947 | **0.9500** | 0.8636 | **0.9211** | 0.0000 |
| LLaMA-8b | 0-shot | 0.6184 | 0.6298 | 0.6753 | 0.6552 | 0.2368 |
| | 1-shot | 0.6955 | 0.6892 | 0.7071 | 0.7368 | 0.0000 |
| | 3-shot | 0.7738 | 0.7703 | 0.8249 | 0.7895 | 0.0000 |
| | 5-shot | 0.7989 | 0.7898 | 0.8434 | 0.8158 | 0.0000 |
| MedGemma-27b | 0-shot | 0.7500 | 0.7415 | 0.8036 | 0.7778 | **0.2895** |
| | 1-shot | 0.7286 | 0.7471 | 0.7172 | 0.7895 | 0.0000 |
| | 3-shot | **0.9013** | 0.9143 | **0.8906** | **0.9211** | 0.0000 |
| | 5-shot | **0.9013** | 0.9143 | **0.8906** | **0.9211** | 0.0000 |

**Table 6.** Subscale Performance on Disclosure Concerns

| Model | Methods | Macro F1 | Macro Precision | Macro Recall | Accuracy | Failure Rate |
|---|---|---|---|---|---|---|
| **GT-large** | SFT | 0.7841 | 0.7841 | 0.7841 | 0.7895 | 0.0000 |
| BERT | SFT | 0.7841 | 0.7841 | 0.7841 | 0.7895 | 0.0000 |
| GPT-OSS-20b | 0-shot | 0.7491 | 0.7646 | 0.7443 | 0.7632 | 0.0000 |
| | 1-shot | **0.8246** | **0.8929** | **0.8125** | **0.8421** | 0.0000 |
| | 3-shot | 0.7564 | 0.8667 | 0.7500 | 0.7895 | 0.0000 |
| | 5-shot | 0.7564 | 0.8667 | 0.7500 | 0.7895 | 0.0000 |
| LLaMA-8b | 0-shot | 0.6667 | 0.6696 | 0.6696 | 0.6667 | 0.2105 |
| | 1-shot | 0.6571 | 0.7673 | 0.6648 | 0.7105 | 0.0000 |
| | 3-shot | 0.7318 | 0.8065 | 0.7273 | 0.7632 | 0.0000 |
| | 5-shot | 0.7662 | 0.8250 | 0.7585 | 0.7895 | 0.0000 |
| MedGemma-27b | 0-shot | 0.7679 | 0.7937 | 0.7798 | 0.7692 | **0.3158** |
| | 1-shot | 0.6381 | 0.8333 | 0.6562 | 0.7105 | 0.0000 |
| | 3-shot | 0.6801 | 0.8438 | 0.6875 | 0.7368 | 0.0000 |
| | 5-shot | 0.7318 | 0.8065 | 0.7273 | 0.7632 | 0.0000 |

**Table 7.** Subscale Performance on Personalized Stigma

| **Model** | **Methods** | **Macro F1** | **Macro Precision** | **Macro Recall** | **Accuracy** | **Failure Rate** |
|---|---|---|---|---|---|---|
| **GT-large** | SFT | 0.6162 | 0.6250 | 0.6122 | **0.6842** | 0.0000 |
| Bert | SFT | 0.4062 | 0.3421 | 0.5000 | **0.6842** | 0.0000 |
| GPT-OSS-20b | 0-shot | 0.3588 | 0.3739 | 0.3558 | 0.3947 | 0.0000 |
| | 1-shot | 0.5117 | 0.5145 | 0.5160 | 0.5526 | 0.0000 |
| | 3-shot | 0.5529 | 0.5523 | 0.5545 | 0.6053 | 0.0000 |
| | 5-shot | 0.5908 | 0.5893 | 0.5962 | 0.6316 | 0.0000 |
| LLaMA-8b | 0-shot | 0.3324 | 0.4107 | 0.4178 | 0.3333 | 0.2895 |
| | 1-shot | 0.5692 | 0.5696 | 0.5769 | 0.6053 | 0.0000 |
| | 3-shot | 0.4824 | 0.5000 | 0.5000 | 0.5000 | 0.0000 |
| | 5-shot | 0.5142 | 0.5361 | 0.5417 | 0.5263 | 0.0000 |
| MedGemma-27b | 0-shot | **0.6406** | **0.6429** | **0.6569** | 0.6538 | **0.3158** |
| | 1-shot | 0.4337 | 0.4354 | 0.4327 | 0.5000 | 0.0000 |
| | 3-shot | 0.5369 | 0.5526 | 0.5609 | 0.5526 | 0.0000 |
| | 5-shot | 0.5447 | 0.5728 | 0.5833 | 0.5526 | 0.0000 |

**Discussion and Conclusions**

In this study, we developed an LLM-based NLP tool to extract and determine HIV stigma from clinical notes, facilitating the study of stigma in HIV treatment outcomes. We also conducted a systematic evaluation across fine-tuned encoder models and prompt-based generative large language models. Using expert-annotated clinical narratives grounded in established HIV stigma theory, we demonstrate that stigma-related language, despite being implicit, heterogeneous, and clinically nuanced, can be detected with meaningful accuracy using LLMs. This work establishes a foundation for scalable measurement of HIV stigma in real-world clinical data and offers concrete guidance for future research and responsible clinical deployment.

Our findings highlight the challenges of HIV stigma extraction and determination. Stigma is often documented indirectly through affective descriptions, social behaviors, or clinician interpretations. This complexity is reflected in a very low inter-annotator agreement (Entity kappa is only 0.286), indicating that HIV stigma detection is a 'Challenge' task even for human experts. Our NLP tool achieved the best F1 score of 0.6190, which is reasonable given the low agreement among human experts.

Among the compared LLMs, our HIV stigma model, fine-tuned with domain-adapted encoder models, delivers the best performance. Another contribution of this work is the demonstration that few-shot generative large language models can closely approach or outperform fine-tuned encoder performance, particularly on linguistically explicit subscales such as 'Negative Self-Image'. These results support the emerging paradigm that text-to-text generative modeling can serve as a cost-effective alternative to task-specific fine-tuning [23,24]. At the same time, our analysis underscores important considerations regarding the reliability and safety of generative LLMs. Zero-shot generative inference is associated with non-trivial failure rates, including non-responses and format violations, which limit its adoption for HIV stigma studies. The introduction of even minimal in-context supervision (one-shot prompting) eliminates these failures across models, underscoring the importance of prompt design and task grounding for stable clinical NLP applications using generative LLMs.

From a clinical perspective, the automated identification of HIV stigma signals from EHR narratives carries important clinical implications. HIV stigma is closely associated with mental health, engagement in care, and treatment adherence; however, it is undermeasured over time or systematically in routine practice. The NLP tool developed in this study provides a practical AI tool to monitoring stigma-related psychosocial risk factors. This approach enables population-level analyses and has the potential to inform targeted interventions.

Future work will focus on aggregating sentence-level predictions into patient- and cohort-level stigma representations and incorporating modeling of temporal dynamics. Additional directions include external validation across healthcare systems, integration with structured clinical variables, and extension to other stigmatized conditions.

**Acknowledgment**
This study was partially supported by grants from NIAID R01AI172875, the Patient-Centered Outcomes Research Institute® (PCORI®) Award ME-2023C3-35934, the PARADIGM program awarded by the Advanced Research Projects Agency for Health (ARPA-H), National Institute on Aging U24AG098157, National Institute of Allergy and Infectious Diseases, National Heart, Lung, and Blood Institute, R01HL169277, R01HL176844, National Institute on Drug Abuse, NIDA R01DA057886, R01DA063631, and the UF Clinical and Translational Science Institute. The content is solely the responsibility of the authors and does not necessarily represent the official views of the funding institutions.

**Appendix**

**Appendix Table 1. HIV Stigma Annotation Guideline**

| Category | Examples | Definition |
|---|---|---|
| **Personalized Stigma** | • I have been hurt by how people reacted to learning I have HIV.<br>• I have stopped socializing with some people because of their reactions to my having HIV.<br>• I have lost friends by telling them I have HIV. | Direct personal experiences of discrimination, prejudice, or negative treatment that an individual faces due to a stigmatized condition. |
| **Disclosure Concern** | • I am very careful who I tell that I have HIV.<br>• I worry that people who know I have HIV will tell others. | The act of revealing or sharing one's stigmatized condition or identity with others. |
| **Negative Self-image** | • I feel that I am not as good a person as others because I have HIV.<br>• Having HIV makes me feel unclean.<br>• Having HIV makes me feel that I'm a bad person. | The internalized feelings of shame, guilt, or low self-worth that an individual develops due to their stigmatized condition, characteristic, or identity. |
| **Concern with Public Attitudes** | • Most people think that a person with HIV is disgusting.<br>• Most people with HIV are rejected when others find out. | Collective beliefs, opinions, and perceptions held by society about a stigmatized condition. |